\documentclass[letterpaper, 10 pt, conference]{ieeeconf}
\IEEEoverridecommandlockouts
\usepackage{cite}
\usepackage{amsmath,amssymb,amsfonts}
\usepackage{hyperref} 
\usepackage{algorithm}
\usepackage{algpseudocode}
\usepackage{graphicx}
\usepackage{textcomp}
\usepackage{xcolor}
\def\BibTeX{{\rm B\kern-.05em{\sc i\kern-.025em b}\kern-.08em
    T\kern-.1667em\lower.7ex\hbox{E}\kern-.125emX}}
\begin{document}



\title{VITaL Pretraining: Visuo-Tactile Pretraining for Tactile and Non-Tactile Manipulation Policies}
\author{Abraham George$^{1}$, Selam Gano$^{1}$, Pranav Katragadda$^{1}$, and Amir Barati Farimani$^{1}$
\thanks{$^{1}$With the Department of Mechanical Engineering,
        Carnegie Mellon University 
        {\tt\small \{aigeorge, selamg, pkatraga, afariman\} @andrew.cmu.edu}}%
}
\maketitle

\begin{abstract}

Tactile information is a critical tool for dexterous manipulation. As humans, we rely heavily on tactile information to understand objects in our environments and how to interact with them. We use touch not only to perform manipulation tasks but also to learn how to perform these tasks. Therefore, to create robotic agents that can learn to complete manipulation tasks at a human or super-human level of performance, we need to properly incorporate tactile information into both skill execution and skill learning. In this paper, we investigate how we can incorporate tactile information into imitation learning platforms to improve performance on manipulation tasks. We show that incorporating visuo-tactile pretraining improves imitation learning performance, not only for tactile agents (policies that use tactile information at inference), but also for non-tactile agents (policies that do not use tactile information at inference). For these non-tactile agents, pretraining with tactile information significantly improved performance (for example, improving the accuracy on USB plugging from 20\% to 85\%), reaching a level on par with visuo-tactile agents, and even surpassing them in some cases. For demonstration videos and access to our codebase, see the project website: \href{https://sites.google.com/andrew.cmu.edu/visuo-tactile-pretraining}{https://sites.google.com/andrew.cmu.edu/visuo-tactile-pretraining}
\end{abstract}

\section{Introduction}
Achieving proficiency in complex manipulation tasks remains a longstanding challenge in robotics, with applications ranging from industrial automation to clay sculpting \cite{stenmark2015knowledge, bartsch2023sculptbot}. Critical to addressing this challenge is the integration of tactile information, which provides both an understanding of the objects being interacted with and a detailed feedback response for closed-loop control. As humans, we instinctively rely on tactile feedback to navigate and manipulate objects with precision, leveraging sensory cues to modulate grip force, detect surface textures, and discern subtle changes in object properties \cite{tactileSensoryControl}. In addition to using this tactile information to complete tasks, we also rely heavily on our tactile perception to learn manipulation skills \cite{6824249}.

\begin{figure}[thpb]
      \centering
      \includegraphics[width=0.95\linewidth]{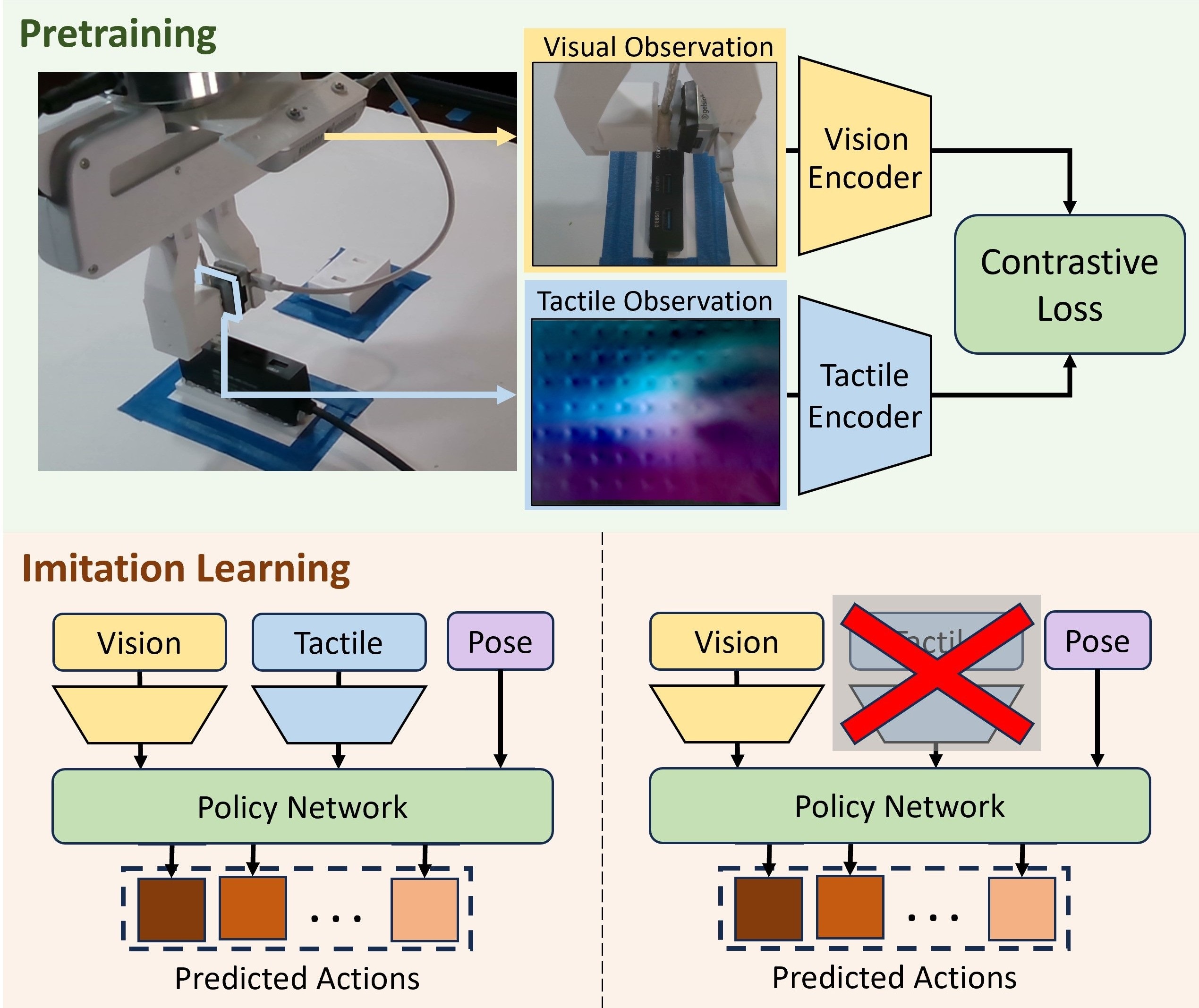}
      \vspace{-5pt}
      \caption{\label{fig:overview} Diagram of our approach. First, a vision encoder and a tactile encoder are pretrained on the collected demonstrations using a temporally informed multi-modal contrastive loss. Then, the pretrained encoders are used in an imitation learning framework, either for visuo-tactile control (left) or vision-only control (right).}
      \vspace{-15pt}
      \label{fig:overview}
\end{figure}
In the realm of robotics, replicating this nuanced interplay between touch and manipulation is an ongoing field of research. Recent advancements in tactile sensing technologies, pioneered by devices such as GelSight \cite{yuan2017gelsight}, have enabled the incorporation of complex tactile information in robotic control pipelines. Prior works have used this tactile information to complete a wide range of manipulation tasks, such as cable routing, object detection, peg insertion, and door opening \cite{she2021cable, kaboli2019tactile, guzey2023dexterity, ablett2023push}. In these works, tactile information was incorporated into multiple control policies, including classical control, reinforcement learning, and imitation learning. 

In this paper, we build upon these prior works, integrating tactile information into two imitation learning frameworks: Action Chunking Transformers (ACT) \cite{zhao2023learning} and Diffusion Policy \cite{chi2023diffusion}. Our primary contribution is a pretraining strategy for these SOTA imitation learning frameworks, leveraging the multimodal nature of our data to incorporate a temporal-based visual-tactile contrastive-loss pretraining step. 


In this step, a tactile encoder and an image encoder are trained to project their respective input modalities onto a shared latent space, with the goal of producing similar embeddings from multi-modal observations of the same scene. This task forces the encoders to learn the relationship between visual and tactile observations and to emphasize features present in both modalities - primarily contact-related features - which are important for dexterous manipulation. First, we use these pretrained encoders as the observation backbone in a visuo-tactile imitation learning system, allowing the agent to better leverage existing multi-modal relationships within its visuo-tactile training dataset. Next, we propose a new methodology for using tactile data in imitation learning: VITaL (\textbf{V}ison-only \textbf{I}mitation using \textbf{Ta}ctile \textbf{L}atent) pretraining, in which we discard the tactile encoder and use the pretrained vision encoder as the backbone for a vision-only imitation learning framework. By pretraining a vision-only policy with the multi-modal dataset, the policy can leverage an implicit tactile understanding without requiring tactile information during deployment. 

To evaluate our visuo-tactile imitation learning framework, we trained ACT and Diffusion Policy agents to plug in a USB cable - a challenging high-precision variant of peg insertion that relies heavily on tactile information, along with two block-stacking tasks. Our experiments show that multi-modal pretraining moderately improves the performance of visuo-tactile imitation learning policies, and significantly improves the performance of vision-only policies, allowing them to achieve success rates comparable to visuo-tactile agents.


\section{Related Works}

\subsection{Learning Control Policies using Tactile Sensors}
Many prior works have sought to use tactile sensors to develop robotic control policies, either through classical control, reinforcement learning, or imitation learning. \cite{wilson2023cable, palli2019tactile} combined classical control and heuristic approaches to tackle the problem of cable following and inserting wires. \cite{she2021cable} went further, combining an LQR cable tracking controller with a fix-position headphone jack and a heuristic-based plugging controller to plug in a headphone cable. Moving beyond hardcoded policies,  \cite{sferrazza2023power, chen2022visuo, yang2023seq2seq} combined visual and tactile observations to train RL agents to complete a range of tasks, including peg insertion, pick and place, and door opening. 
To avoid the large amount of data RL training requires ($\sim$1 million steps), multiple works have used Nearest Neighbor Imitation Learning \cite{pari2021surprising}, which mimics the action of the demonstration in its training data whose corresponding observation is nearest to the current observation, measured via distance in an encoded latent space. \cite{yu2023mimictouch} used this strategy, in combination with an RL residual policy, to learn visuo-tactile peg-insertion, and \cite{guzey2023dexterity} used it to learn a variety of manipulation tasks including bowl unstacking, book opening, and joystick movement. Although VINN shows good performance with surprisingly little data, it cannot generalize, causing it to under-perform direct action-prediction policies such as Action Chunking Transformers (ACT) \cite{zhao2023learning}.

\subsection{Contrastive Pretraining}
Contrastive pretraining is a strategy to learn relevant features with a dataset by training a model to associate like pairs of data and disassociate unlike pairs, usually by constructing a shared latent space \cite{rethmeier2023primer}. This type of pretraining has been used in a wide range of fields from image classification \cite{chen2020simple}, to symbolic regression \cite{meidani2023snip}, to molecular property prediction \cite{wang2022molecular}, to modeling partial differential equations \cite{lorsung2024picl}. Contrastive pretraining is especially useful for multi-modal datasets, where different modalities of the same observation, such as an image and a text caption of that image, can be paired \cite{radford2021learning}. In the field of visuo-tactile robotic systems, contrastive pretraining has been used to directly link vision and tactile observations, for example in identifying garment flaws \cite{kerr2022self}, and improving few-shot learning of visuo-tactile object classification \cite{zambelli2021learning, dave2024multimodal}. Additionally, this method has been applied to control tasks, with \cite{10610933} incorporating visuo-tactile pretraining into reinforcement learning to train a robotic hand to put a cap on a bottle (trained in sim). However, the literature is lacking in visuo-tactile pretraining for non-VINN imitation learning, and no prior works (to our knowledge) have looked into visuo-tactile pretraining for non-tactile control policies.


\subsection{Imitation Learning for Robotic Manipulation}
Imitation learning casts the control problem as a supervised learning task: given a set of observation action pairs, train an agent to fit goal actions to given observations \cite{pomerleau1988alvinn}. This method of control has been used for tasks ranging from automobile operation \cite{ahn2022autonomous}, to biped locomotion \cite{nakanishi2004}, to robotic manipulation tasks from block stacking \cite{george2023one} to kitchen chores \cite{shafiullah2022}. However, imitation learning struggles with complex action spaces, non-deterministic goal policies, and significant shifts between training and testing (deployment) distributions \cite{hussein2017imitation}. Multiple solutions have been proposed to address these concerns, but we focus on two: Action Chunking Transformers (ACT) which use temporal ensembling to limit the effect of out-of-distribution actions, and Diffusion Policy, which uses diffusion to model complex action distributions. 

\subsubsection{Action Chunking Transformers}
Action Chunking Transformers (ACT) \cite{zhao2023learning} train a Conditional Variational Auto Encoder (CVAE) built upon a transformer backbone to predict a series of actions (in the form of goal positions) conditioned on state observations (position + vision). The use of an auto-encoder for this task helps to reduce the negative effects of multi-modal distributions in the training data, as the latent variable can encode the "style" of the goal action sequence. At inference, the latent variable is set to the mean of the prior (zero) to generate the most likely action. A strong KL divergence term in the loss function encourages the network to avoid over-reliance on the latent variable.

Because ACT predicts a sequence of actions (goal states), $A_{t,t}$ to a $A_{t,t+h}$ at each timestep, during inference a series of previous predictions $A_{t-h, t}$ to $A_{t-1, t}$ is available. To reduce the effect of a single bad prediction (often caused by out-of-distribution states), this series of predictions is combined using a weighted average with exponentially decaying weights, $w_i = e^{-ki}$, and the resulting ensembled action is executed.

\subsubsection{Diffusion Policy}

To address the challenge of complex multi-modal action spaces, Diffusion Policy \cite{chi2023diffusion} formulates control policies as Denoising Diffusion Probabilistic Models (DDPM)\cite{ho2020denoising}, generating action sequences conditioned on observations, $p(\mathbf{O}_t|\mathbf{A}_t)$, through iterative denoising of the action sequence:

\vspace{-10pt}
$$\mathbf{A}_t^{k-1} = \alpha(\mathbf{A}_t^k- \gamma \epsilon_\theta(\mathbf{O}_t,\mathbf{A}_t^k,k)+\mathcal{N}(0,\sigma^2I))$$

where $\epsilon_\theta$ is a noise prediction network with learned parameters $\theta$, and $\alpha$, $\gamma$, and $\sigma$ are parameters representing the noise schedule. 




The observation $\mathbf{O}_t$ is used to condition the noise prediction via Feature-Wise Linear Modulation (FiLM) layers in the noise prediction network \cite{perez2018film, chi2023diffusion}. This network is trained to predict noise added to action sequence samples from the training dataset, minimizing the loss function: 

$$Loss=MSE(\epsilon^k,\epsilon_\theta(\mathbf{O}_t,\mathbf{A_t}^0+\epsilon^k,k))$$

where the variance of random noise $\epsilon^k$ at iteration k is determined by the noise scheduler.







Additionally, \cite{chi2023diffusion} decoupled the training and inference noise scheduler, allowing for fewer denoising iterations during inference to improve speed. 


\section{Methods}
\subsection{Pretraining}
To leverage tactile observations to improve the training of imitation learning agents, we use contrastive pretraining to instill an understanding of the relationships between tactile, visual, and positional observations of the scene. To do this, we pretrain a vision encoder, $f_{\theta_V}: \mathcal{V} \rightarrow \mathcal{Z}$, and a tactile encoder, $g_{\theta_T}: \mathcal{T} \rightarrow \mathcal{Z}$, using a CLIP-inspired contrastive-loss. These encoders, which use the ResNet-18  architecture \cite{he2016deep}, extract a feature vector, $\mathcal{Z} \in \mathbb{R}^{512}$, from each of the visual observations, $\mathcal{V} \in \mathbb{R}^{H_V\times W_V\times3}$, and the tactile observation, $\mathcal{T} \in \mathbb{R}^{H_T\times W_T\times3}$, respectively.

\begin{figure}[thpb]
      \centering
      \includegraphics[width=0.97\linewidth]{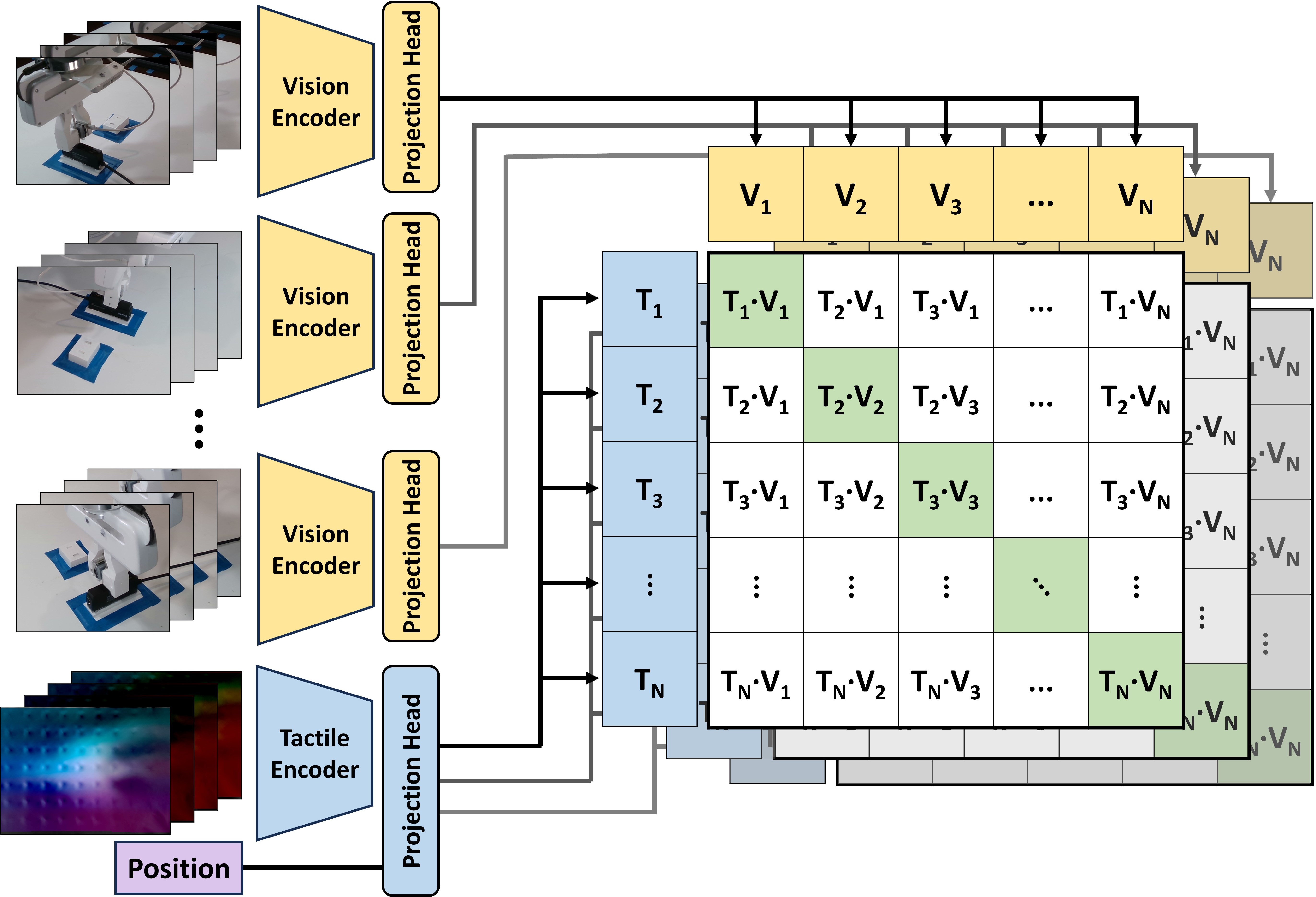}
      \caption{\label{fig:clip_diagram} Contrastive loss visualization. A series of visual observations $V_1, V_2, ..., V_N$ and tactile observations $T_1, T_2, ..., T_N$ are collected, and the vision encoder and tactile encoder are trained to make the embeddings from the same timestep similar while forcing apart the embeddings from different time steps.}
      \label{clip_diagram}
\end{figure}

A vision projection head, $p_{\phi_V}: \mathcal{Z} \rightarrow \mathcal{L}$, and a tactile projection head, $q_{\phi_T}: \mathcal{Z}, \mathcal{P} \rightarrow \mathcal{L}$, each of which consists of a single layer MLP, map the encoders' feature vectors to a shared latent space, $\mathcal{L} \in \mathbb{R}^{512}$. In addition to the tactile feature vector, the tactile projection head takes in the current position of the robot, $\mathcal{P} \in \mathbb{R}^{3}$. By combining the tactile and positional information this way, we ensure that the tactile latent representation has access to both local information (from the tactile observation) along with some global information (from the positional data). The encoders (and the projection heads) are then trained to maximize the cross-modality dot-product similarity of latent representations from the same scene while minimizing the cross-modality similarity of latent representations from different scenes. After pretraining, the projection heads are discarded. A diagram illustrating our pretraining strategy can be found in Figure \ref{fig:clip_diagram}, and the detailed implementation can be seen in Algorithm \ref{alg:logic}.

\begin{algorithm}
\caption{\label{alg:logic} Contrastive Pretraining}
\label{figurelabel}
\begin{algorithmic}
\State \textbf{Input:}$\,\{V$\},\,$\{T$\},\,$\{P\}$ \Comment{Vision, Tactile, and Position Obs}

\State \textbf{Initialize:} $f_{\theta_V}, g_{\theta_T}$ 
\Comment{Vision Encoder, Tactile Encoder}

\State \textbf{Initialize:} $p_{\phi_V}, q_{\phi_T}$ 
\Comment{Vision Proj-head, Tactile Proj-head}

\While{Training}
    \State Sample Trajectory $V$, $T$, $P$, from set of observations
    \State Sample $t = \{t_1, t_2, ... t_n\}$, st. $|t_i - t_j| > \Delta t_{min} \; \forall \:i, j$
    \For{$i \in t$}
        \State $L_{T_{i}} = q_{\phi_T}(g_{\theta_T}(T_i), P_i)$
        \For{c in cameras}
            \State $L_{V_{i, c}} = p_{\phi_V}(f_{\theta_V}(V_{i, c}))$
        \EndFor
    \EndFor
    \State $sim_{i, j, c} = ||L_{T_{i}}||\bullet||L_{V_{i, c}}||$ \Comment{Similarity}
    \State $SM1_{i, c} = \frac{exp(sim_{i, i, c}/\tau)}{\sum_{j \in t} exp(sim_{i, j, c}/\tau)}$ \Comment{Softmax of $sim$}
    \State $SM2_{i, c} = \frac{exp(sim_{i, i, c}/\tau)}{\sum_{j \in t} exp(sim_{j, i, c}/\tau)}$ \Comment{Softmax, other axis}
    \State $loss = -\sum_{c}\sum_{i \in t} \frac{1}{2n}(log(SM1_{i, c}) + log(SM2_{i, c}))$
    \State Update $\theta_V, \theta_T, \phi_V, \phi_T$ using $loss$
\EndWhile

\end{algorithmic}
\end{algorithm}


To form the contrastive pairs for the pretraining, we sampled $n$ (we used 7) observations randomly from a single demonstration, with the requirement that the samples must be at least $\Delta t_{min}$ time steps apart (we used 10, which equates to 1 second). By only sampling observation pairs from a single trajectory, and ensuring they are sufficiently separated in time, we instilled previously unused temporal information into the encoder, as the model had to learn how the scene changed over time, rather than in between runs. 

For each timestep, we have multiple visual observations, one from each camera. During training, the contrastive loss between tactile and visual embeddings is calculated separately for each camera observation, and the resulting losses are combined and used to update the network. This strategy helps ensure balanced training of the vision encoder, which is shared across all cameras. Additionally, by enforcing similarity between the tactile embedding and each vision embedding for a given timestep, we implicitly align the embeddings of different views of the same scene. By having the cameras share an aligned latent space, we encourage the vision embedding to learn general features of the scene, rather than camera-specific characteristics.


\subsection{Imitation Learning Frameworks}
\begin{figure}[thpb]
      \centering

      \includegraphics[width=\linewidth]{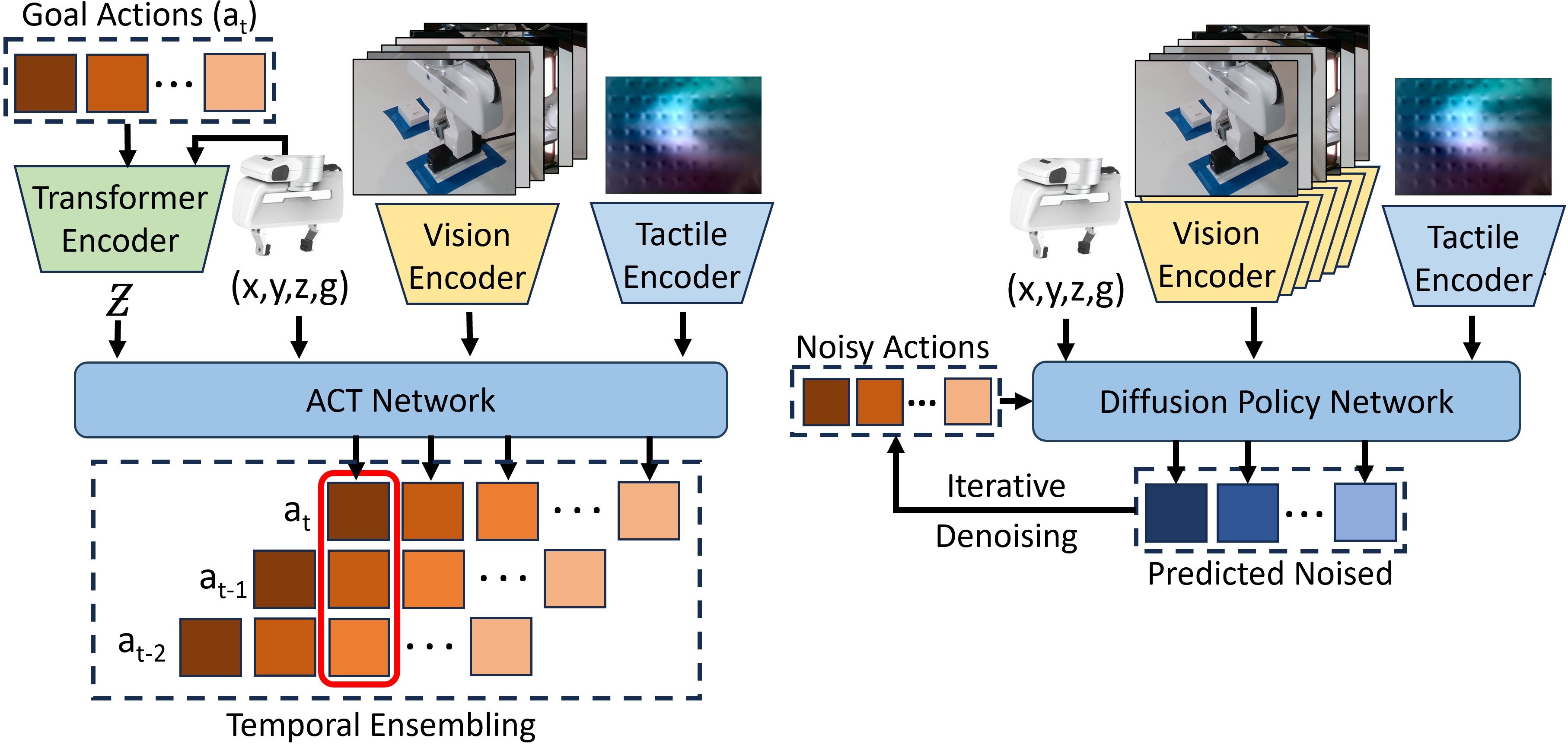}
      \vspace{-15pt}
      \caption{\label{fig:model_diagram} Imitation learning networks. ACT (left) is trained as an autoencoder, predicting a sequence of actions at each timestep ($a_t$). At inference, the latent variable, $z$, is set to 0. The network is queried each timestep, and all action predictions for that timestep are ensembled using a weighted average. Diffusion Policy (right) learns to predict noise applied to an action sequence. During inference, the action sequence is initialized with Gaussian noise and is iteratively denoised to produce output actions.}
      \vspace{-10pt}
      \label{ACT_diagram}
\end{figure}

We implemented two imitation learning frameworks to study the impact of visuo-tactile pretraining on learning manipulation tasks: Action Chunking Transformer (ACT) \cite{zhao2023learning} and Diffusion Policy \cite{chi2023diffusion}. A diagram of our two implementations can be seen in Figure \ref{fig:model_diagram}.

\subsubsection{Action Chunking Transformer}
For our implementation, we made two modifications to the ACT framework. First, we replaced the stock Resnet vision encoder with the vision encoder from the contrastive pretraining step and added a separate tactile encoder (also from the pretraining step) to encode the tactile observations. The tactile encoding is concatenated with the vision encodings before being passed into ACT's transformer decoder. Second, instead of training the network to predict goal positions, we trained it to predict relative goal positions ($goal\_pos_{t:t+h} - pos_{t}$ instead of $goal\_pos_{t:t+h}$). At inference, we add the current position to the predictions to get goal positions in the global coordinate frame. Operating in the delta position space, instead of the position space, allows the network to avoid biases that would be detrimental to the task of visual-servoing. At inference, we used standard temporal ensembling with a temperature constant of $k=0.25$.

\subsubsection{Diffusion Policy}
Our approach to diffusion policy was based on the implementation by \cite{chi2023diffusion} that generates action sequences conditioned on observations with DDPM. We use the CNN-based implementation of this model, where a 1D temporal CNN models the conditional action distribution. For this approach, each observation (visual, tactile, and positional) is passed through an encoder, and the resulting embeddings are stacked to form a single observation vector. For vision observations, we replace the stock ResNet18 encoder with our pre-trained encoder (with separately fine-tuned, identically initialized encoders for each camera view). For tactile observations, we use our pre-trained tactile encoder. The network then uses the observation embedding vector to condition its noise prediction.


We use an observation horizon of 1 and an action prediction horizon of 20. We take advantage of noise scheduler decoupling, using 100 denoising steps during training and 10 steps at inference. At inference, we execute 8 of the 20 predicted actions before replaning. 

\subsection{Data Collection}
The expert demonstrations used to train our imitation learning policies were collected via teleoperation of a Franka Emika Panda robot. We used the Oculus Virtual Reality (VR) teleoperation pipeline developed by \cite{george2023openvr}, which tracks the Quest's controller using the headset. However, we did not use VR for observing the tasks, instead having the operators directly look at the workspace. The teleoperation system provides a goal state (position and gripper width of the robot's end effector) to an impedance controller, which controls the robot. 

\begin{figure}[thpb]
      \centering
      \includegraphics[width=0.97\linewidth]{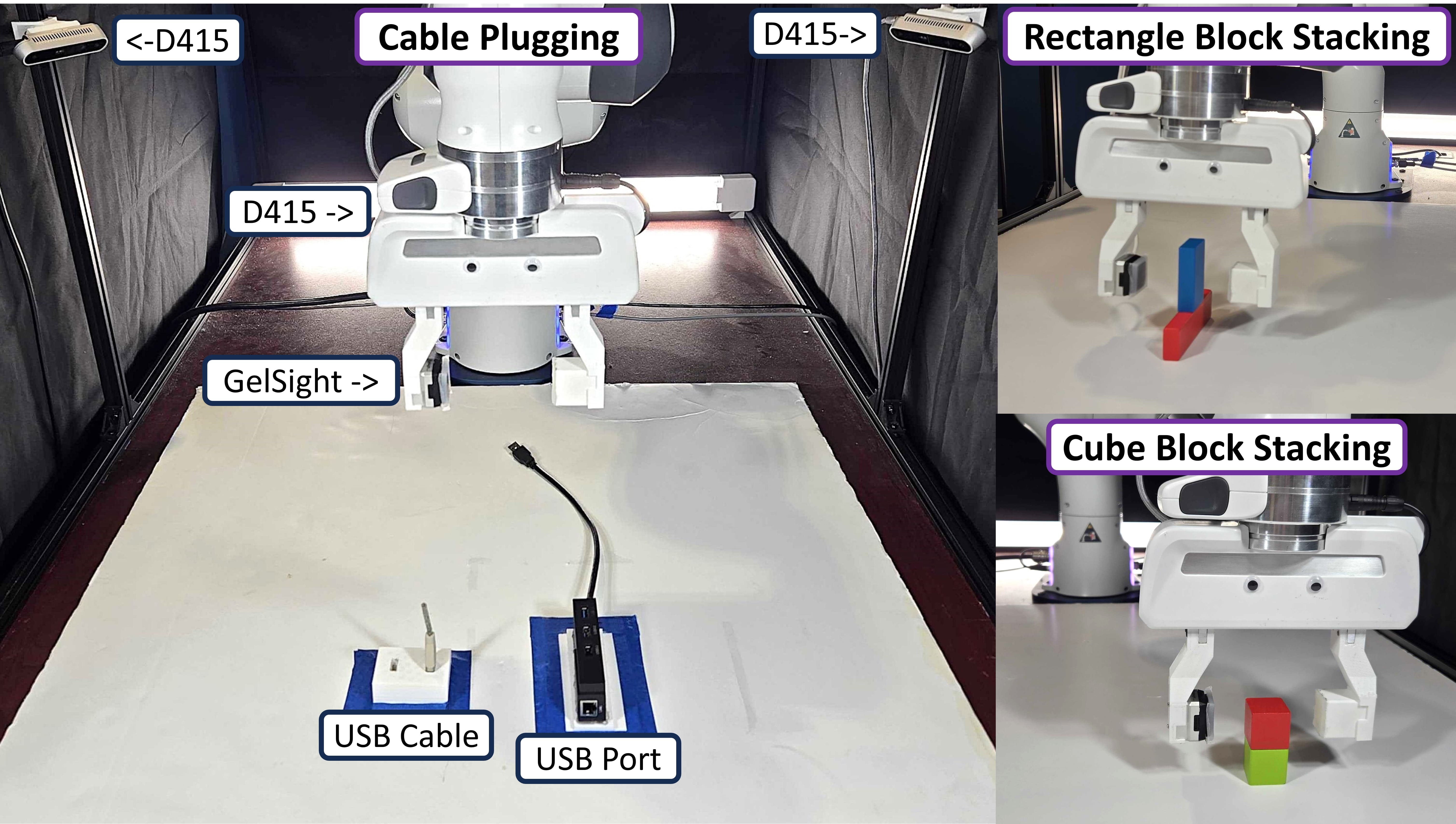}
      \caption{\label{fig:scene} Expermental Setup.  A GelSight captures tactile observations, while 6 Realsense cameras observe the scene (only two can be seen above; three are out of view and another is mounted to the back of the end-effector). Cable Plugging Task (left): The robot must retrieve the USB cable from its holder, plug it into the front port on the USB hub, and release it. Rectangle Block Stacking (top-right): The robot must pick up the blue block, stack it on top of the red block, and release it without knocking over the blocks. Cube Block Stacking (bottom-right): The robot must pick up the red block, stack it on the green block, and release it.}
      \label{fig:gelsight}
\end{figure}

During teleoperation, six RGB views of the workspace were recorded using Realsense cameras, consisting of a D415 wrist-mounted camera, four D415 cameras mounted around the workspace, and a D445 camera mounted on the top edge of the workspace, which provided a scene overview. Additionally, the current state of the robot's end-effector (x position, y position, z position, and gripper width) and the goal state (as measured by the teleoperation system) were recorded. The data collection system was run at 10 Hz.  A view of the workspace can be seen in Figure \ref{fig:scene}.

Tactile observations were provided by a GelSight Mini, which was mounted to the robot's end-effector. During our initial experimentation, we found that the high contact forces involved in cable plugging, along with the repetitive nature of data collection, led to significant wear on the soft silicon gel pad of the GelSight. A GelSight pad was only able to perform around 30 teleoperated cable plugging episodes before tearing. To address this issue, we equipped the GelSight with a 1 mm thick protective Ecoflex 00-50 silicon rubber topper. Although this topper reduced the GelSight's ability to detect small-scale surface features, it preserved the GelSight's perception of large-scale features (contact pressure, contact area, strain, etc.) relevant to robotic manipulation tasks. A side-by-side comparison of the GelSight data with and without the protective cover can be seen in Figure \ref{fig:gelsight}. Despite the use of the cover, two additional GelSight pads were damaged during experimentation. However, the cover managed to extend the lifetime of the GelSight pad by an order of magnitude, withstanding a few hundred plugging episodes before tearing. 


\begin{figure}[thpb]
      \centering
      \includegraphics[width=\linewidth]{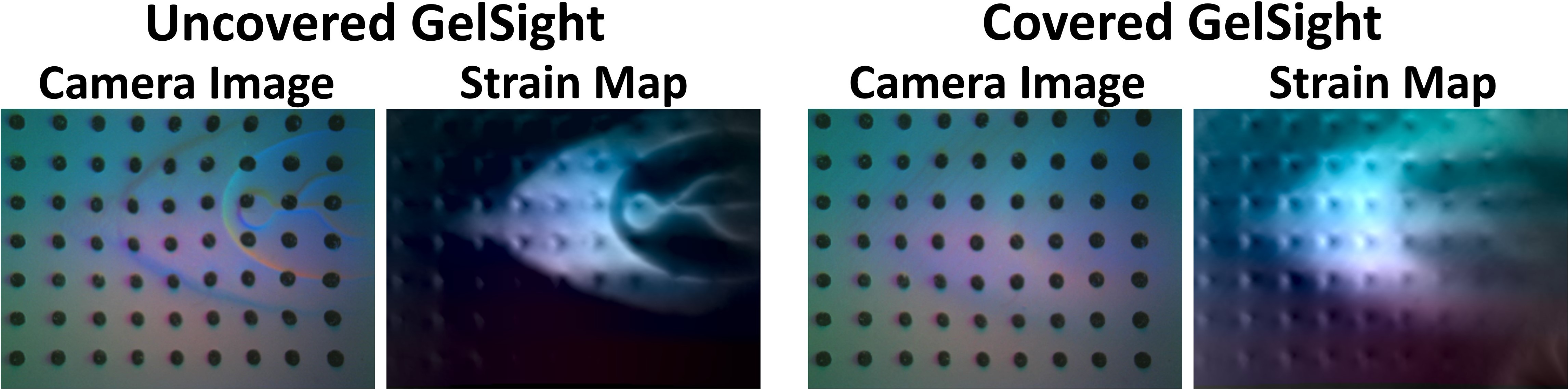}
      \vspace{-15pt}
      \caption{\label{fig:gelsight} GelSight sensor outputs, showing the RGB images from the GelSight's camera and the processed strain data for both the covered and uncovered gelsight. The strain map is rendered in the LAB color space, with the brightness of each pixel corresponding to the normal strain (depth), and the color corresponding to the tangential strains, with strains in x shown on the blue-yellow spectrum, and strains in y shown on the red-green spectrum.}
      \label{fig:gelsight}
      \vspace{-10pt}
\end{figure}

Previous works using GelSight as a control policy input found that using processed data, such as strain information, out-performed directly using RGB images \cite{dong2021tactile}. Based on these observations, and to reduce the impact of discrepancies between different GelSight pads, we used a processed tactile input for our imitation learning pipeline, representing the tactile information as a 240 x 320 map of the gel's strain in the x, y, and z directions. A visualization of this strain data can be found in Figure \ref{fig:gelsight}.

\section{Experimental Evaluation}
\subsection{Cable Plugging}
We first evaluated our model on the task of cable plugging. In this task, the robot has to navigate to a USB cable, unplug it from its holder, and plug it into the last port of a USB hub. We collected 100 demonstrations, with an average length of 208 time steps (approximately 21 seconds). Because we are most interested in the visuo-tactile task of inserting the USB cable, not the vision-only task of localizing the plug and cord, we fixed the start positions of the cord holder and USB plug. To increase the task's difficulty, we added random noise with a standard deviation of 2.5mm to the agent's actions during inference. 

\begin{figure}[thpb]
      \centering
      \includegraphics[width=\linewidth]{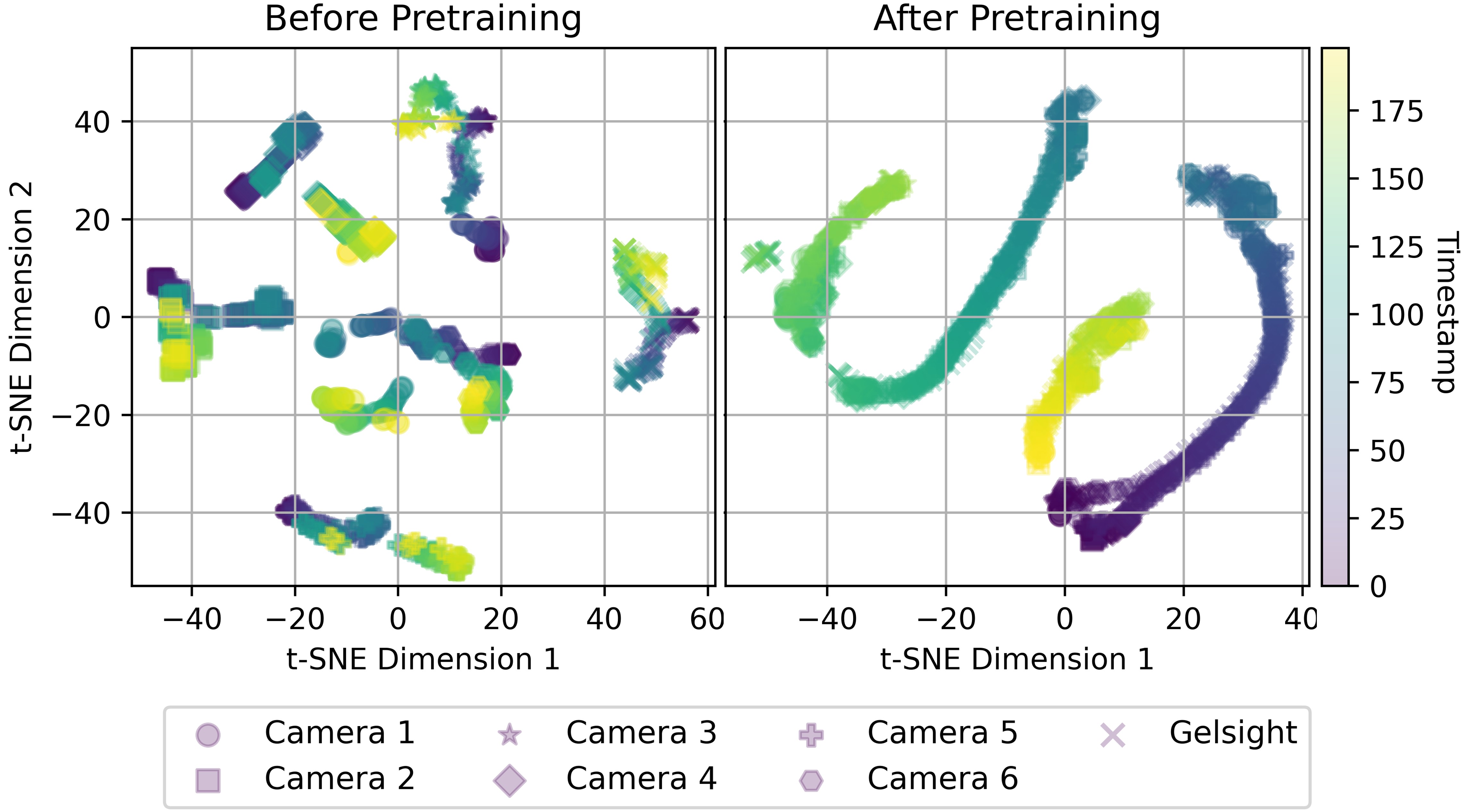}
      \vspace{-10pt}
      \caption{\label{fig:tsne_plot} t-SNE plot comparing the latent space of a demonstration (from the cable plugging testing dataset) before and after visuo-tactile pretraining, showing how pretraining aligns the visual and tactile latent spaces. The two clusters in the after-pretraining plot (right) correspond to timesteps when the robot was not grasping anything (bottom-right cluster) and timesteps when the robot had grasped the cable (top-left cluster). }
      \label{fixed_results}
      \vspace{-15pt}
\end{figure}

Once we had collected the data, we pretrained the visual and tactile encoders, using an 80/20 train/test split. Before using these encoders in the downstream imitation learning frameworks, we evaluated how their latent spaces changed during pretraining using a t-SNE plot (shown in Fig. \ref{fig:tsne_plot}). We observed that pretraining successfully aligned the latent spaces, with the visual and tactile observations' embeddings overlapping after pretraining. Additionally, the pretrained embeddings are aligned by timestep and bisect themselves into two clusters, based on whether the cable was grasped.

We then trained the agents on 80 demonstrations, reserving 20 for validation during the training process.  Durring evaluation, we let the agents run until they successfully plugged in the cable, reached an un-recoverable state (ie. dropped the USB cable), or exceeded 300 time steps. All of our experiments were run 20 times. Due to wear on the original pad, a different gel pad was used during training and testing.

\begin{figure*}[thpb]
      \centering
      \includegraphics[width=0.95\linewidth]{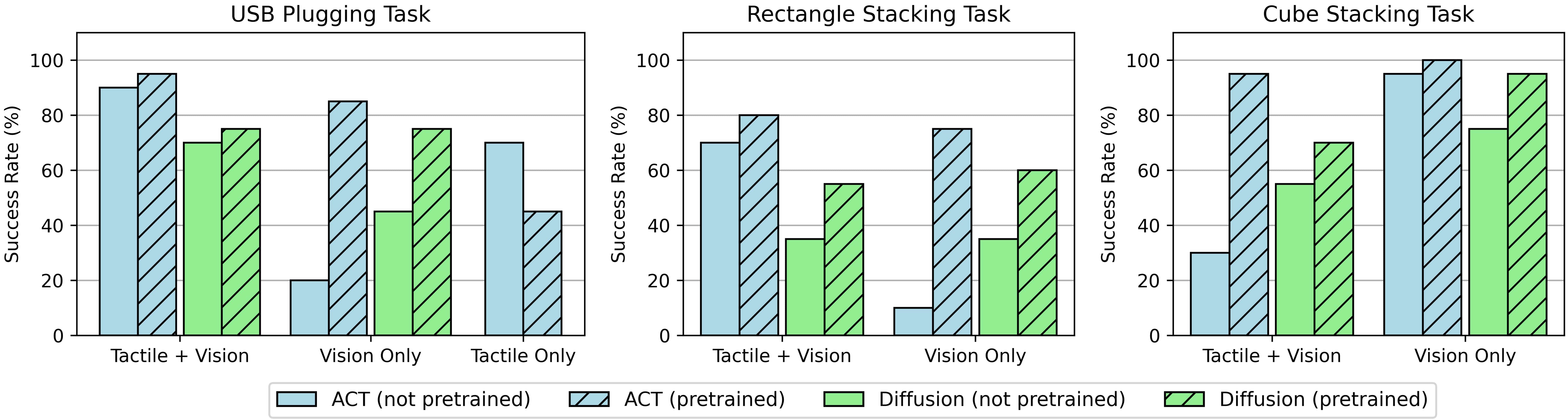}
      \vspace{-5pt}
      \caption{\label{fig:fixed_results} Success rate for our experiment tasks. All experiments were run 20 times, with the total success rate shown. The use of visuo-tactile pretraining (hashed bars) leads to a sight increase in performance for visuo-tactile policies and a significant improvement for vision-only policies.}
      \label{fixed_results}
      \vspace{-10pt}
\end{figure*}

We ran 4 experiments for each imitation learning (IL) method, examining the impact of visuo-tactile pretraining on vision-only policies (IL agent with only visual observations) and visuo-tactile policies (IL agent with both tactile and visual observations). Our results can be seen in Figure \ref{fig:fixed_results}.


We found that by pretraining a visuo-tactile agent, we were able to solve the cable plugging task, reaching a 95\% success rate for ACT. This is significantly higher than the 20\% and 45\% success rates that learning from vision only with ACT and diffusion policy (respectively) achieves. Beyond solving the cable plugging task, our results show that contrastive pretraining results in a modest improvement in performance for both visuo-tactile agents, granting both policies a 5\% performance increase over a non-pretrained visuo-tactile agent. Although this is relatively small in absolute terms, it corresponds to a 50\% and 20\% decrease in failures for ACT and Diffusion Policy, respectively. 


More importantly, pretraining with tactile data led to a significant improvement in the vision-only agent, increasing ACT's performance from 20\% to 85\%, and diffusion policy's from 45\% to 75\%, achieving almost the same level of performance as the visuo-tactile agents. In addition to improving overall performance, we found that pretraining with tactile information decreased the force exerted by the agent. We measured the mean absolute tangential strain at each timestep during execution, and examined the median across all runs, finding that pretraining reduced the strain in visuo-tactile policies by 8\% and 13\%, and by 20\% and 17\% in vision-only policies for ACT and Diffusion Policy, respectively.

Comparing the two imitation learning methods, we found that Diffusion Policy's success rate was less sensitive than ACT, with a higher accuracy for the non-pretrained vision-only policy (where ACT did quite poorly), but lower accuracy for the pretrained vision and tactile+vision results (where ACT did quite well). Diffusion's longer re-plan horizon (re-planning every 8 steps instead of every step) and stochastic nature lead to larger motions, tending to jump around the port before plugging in the USB. These larger more random motions helped the policy avoid getting stuck (like ACT did) when poorly trained but impeded the well-trained policy. Also, the large more stochastic motions of Diffusion Policy caused an increase in contact force, with Diffusion Policy increasing GelSight strain by about 15\% as compared to ACT for the pretrained visuo-tactile policy.

Finally, we evaluated the models without vision input (only tactile and positional data). Interestingly, the non-pretrained ACT model outperformed the pretrained model in this task. We think this may be due to overfitting, especially since we used different GelSight pads for training and evaluation. For both pretrained and non-pretrained tactile-only diffusion policies, the robot would pick up the USB cable, but then would alternate between trying to move back to the USB holder, trying to move towards the USB hub, and opening its gripper to re-grasp the USB (resulting in dropping the USB), never successfully completing the task.

\subsection{Block Stacking}
In addition to cable plugging, we also evaluated our pretraining strategy on two block-stacking tasks to see how well the system performed on more general tasks. The first block-stacking task required the agent to stack a 1.5cm x 3cm x 6cm vertical block onto a 1.5cm x 3cm x 9cm horizontal block (see Fig. \ref{fig:scene}). Stacking these thin rectangular blocks required precise motion and force control to ensure the lower block isn't knocked over and the top block doesn't fall when the gripper opens. The second block stacking task used two 3cm cubes. This task, which doesn't require detailed force/position feedback, was included to see how the system handles an inherently non-tactile task. We used the same run parameters as the cable plugging task, with 100 demos collected, an 80/20 train/test split, and noise added to the predicted actions. Additionally, we randomized the start locations of the blocks, although this ment we were unable to test a vision-only policy for these tasks. The results for these experiments can be seen in Fig. \ref{fig:fixed_results}.

In all block stacking experiments, as with cable plugging, we observed that visuo-tactile pretraining improved performance for both visuo-tactile and vision-only agents, and that a vision-only agent, pretrained with tactile information, was able to achieve results on par with a visuo-tactile agent. Interestingly, unlike the USB plugging task, the visuo-tactile agents did not consistently outperform their vision-only counterparts. Especially in the simple cube block stacking task, the inclusion of tactile data appears to have significantly decreased performance, especially if the agent was not pretrained. This unusual behavior, more data leading to lower performance, is likely do to overfitting. Because the GelSight is very susceptible to wear, the tactile sensor's properties during data collection (the same dataset is used for both pretraining and training) and evaluation will be different, causing a domain shift and resulting in overfitting. This result illustrates the key benefit of using visuo-tactile pretraining on a vision-only agent: the agent gains a significant performance boost from tactile data without the many challenges of deploying a tactile sensor during inference, including sensor wear and overfitting. 

\section{Conclusions}


In this work, we have shown that incorporating contrastive visuo-tactile pretraining into imitation learning frameworks can boost agent performance in complex manipulation tasks. For visuo-tactile agents, this pretraining step grants a moderate performance boost (for example, improving the cable plugging success rate from 90\% to 95\%), and reduces overfitting (as shown in the cube stacking task, where pretraining improved the visuo-tactile success rate from 30\% to 95\%). Additionally, we found that pretraining decreased the force exerted by the agent, reducing wear on the tactile sensor and suggesting that pretraining leads to a contact-aware policy. 

When this pretraining step was applied to vision-only agents, we found that pretraining with tactile data significantly improved performance, allowing these agents to achieve a success rate on par with, and occasionally surpassing, their visuo-tactile counterparts.  For example, pretraining increased the vision-only cable plugging success rate from 20\% to 85\%. These results suggest that visuo-tactile pretraining of a vision-only policy, rather then developing visuo-tactile policies, may be a better use of tactile data in robotics. This training paradigm could be especially useful in industrial applications, where durability and cost concerns limit the use of high-quality tactile sensors like GelSight, allowing imitation learning policies to leverage tactile cues to significantly improve performance without requiring the deployment of tactile sensors.

A major limitation of this work is that task-specific data was used for pretraining. Although this strategy allowed the vision-only policy to learn the tactile features relevant to the specific task, it limits the applicability of the pretrained model and requires tactile observations to be collected for each task. An alternative approach would be to collect a large-scale visuo-tactile dataset for pretraining, then fine-tune a vision-only policy on a separate task, trading task-specific tactile knowledge for increased generalizability. Evaluating this alternative approach is left for future work. 

\bibliographystyle{IEEEtran}
\bibliography{references}
\vspace{12pt}

\end{document}